\pdfoutput=1

\documentclass[11pt]{article}

\usepackage[]{acl}

\usepackage{times}
\usepackage{latexsym}
\usepackage{graphicx}
\usepackage[T1]{fontenc}

\usepackage[utf8]{inputenc}

\usepackage{microtype}
\usepackage{multirow}
\title{bitsa\_nlp@LT-EDI-ACL2022: Leveraging Pretrained Language Models for Detecting Homophobia and Transphobia in Social Media Comments}

\author{Vitthal Bhandari \and Poonam Goyal\\
  Birla Institute of Technology and Science, Pilani, India \\
  \texttt{f20170136p@alumni.bits-pilani.ac.in} \\ \texttt{poonam@pilani.bits-pilani.ac.in} \\}

\begin{document}
\maketitle
\begin{abstract}
Online social networks are ubiquitous and user-friendly. Nevertheless, it is vital to detect and 
moderate offensive content to maintain decency and empathy. However, mining social media texts is a complex task since users don't adhere to any fixed patterns. Comments can be written in any combination of languages and many of them may be low-resource.

In this paper, we present our system for the LT-EDI shared task on detecting homophobia and transphobia in social media comments. We experiment with a number of monolingual and multilingual transformer based models such as mBERT along with a data augmentation technique for tackling class imbalance. Such pretrained large models have recently shown tremendous success on a variety of benchmark tasks in natural language processing. We observe their performance on a carefully annotated, real life dataset of YouTube comments in English as well as Tamil.

Our submission achieved ranks $9$, $6$ and $3$ with a macro-averaged F1-score of $0.42$, $0.64$ and $0.58$ in the English, Tamil and Tamil-English subtasks respectively. The code for the system has been open sourced\footnote{The code for this task is available at \href{https://github.com/vitthal-bhandari/Homophobia-Transphobia-Detection}{github.com/vitthal-bhandari/Homophobia-Transphobia-Detection}.}.
\end{abstract}

\section{Introduction}

Twenty first century social media has become the epicenter of polarized opinions, arguments, and claims. The ease of information access not only benefits fruitful discussions but also facilitates phenomena such as hate speech and cyber bullying. 

Recently organized workshops and shared tasks have fostered discussions around detection of hate speech, toxicity, misogyny, sexism, racism and abusive content \citep{zampieri-etal-2020-semeval, mandl2020overview}. While research in processing and classifying offensive language in social media is vast \citep{pamungkas2021towards}, there is very little work on detecting sexual orientation discrimination in particular.  More so, compared to resource-rich languages such as English and Japanese, Indic languages such as Tamil and Malayalam are scarce in well-annotated data. Although advancements in large multilingual models have promoted cross-lingual transfer learning in Indic languages \cite{dowlagar2021survey}, there have not been any visible attempts to censor homophobia and transphobia. The perception of the subject matter as being taboo prohibits advancements in data collection, annotation and analysis.

Curbing sensitive online content is imperative for preventing harm to mental health of the community as well as avoiding divide between minorities. These reasons have contributed towards the need of moderating social media comments spreading any form of hatred towards the LGBTQIA+ population. 

While both - the detection of homophobia/transphobia and the corresponding research in Indic languages - is underserved and low-resource, another factor contributing to the difficulty in processing social media texts is code-mixing - a phenomena in which multilingual speakers switch between two or more languages in a conversation with the aim to be more expressive. Popular language models tend to perform adversely when applied to code-mixed text and hence newer techniques need to be adopted to handle this situation \citep{dougruoz2021survey}.

The pretraining and fine-tuning paradigm has taken extensive advantage of transformer based large multilingual models which perform well in cross-lingual scenarios. In this paper we explore the performance of a number of such models when fine-tuned on a dataset for detecting homophobia and transphobia. Surprisingly, our experiments also show that these multilingual models exhibit reasonably accurate performance on code-mixing tasks, even without any previous exposure to code-mixing during pretraining.

The remainder of the paper is organized as follows: Section \hyperlink{section.2}{2} talks about the previous related work in this domain. Section \hyperlink{section.3}{3} gives a detailed explanation of the methods used in the system and Section \hyperlink{section.4}{4} describes the corresponding experimental settings. We mention the detailed results in Section \hyperlink{section.5}{5}, conduct an ablation study in Section \hyperlink{section.6}{6} and conclude our discussion with Section \hyperlink{section.7}{7}.

\section{Related Work}

To the best of our knowledge no prior work identifying either homophobia or transphobia directly exists in recent literature. However, offensive language detection, in general, in Dravidian languages has been the focus of multiple research works in the past \citep{chakravarthi-etal-2021-findings-shared, mandl2020overview}.

\citet{baruah2021iiitg} at HASOC-Dravidian-CodeMix-FIRE2020 trained an SVM classifier using TF-IDF features on code-mixed Malayalam  text and an XLM-RoBERTa based classifier on code-mixed Tamil text to detect offensive language in Twitter and YouTube comments. \citet{sai2020siva} fine-tuned multilingual transformer models and used a bagging ensemble strategy to combine predictions on the same task.

\citet{saha-etal-2021-hate} developed fusion models by ensembling CNNs trained on skip-gram word vectors using FastText along with fine-tuned BERT models. A neural classification head was trained on the concatenated output obtained from the ensemble.

A number of approaches have been deployed to tackle code mixing in Indic languages as well, since multilingual transformer models lack the complexity to extract linguistic features directly from code switched text. \citet{Vasantharajan2021} used a selective translation and transliteration technique to process Tamil code-mixed YouTube comments for offensive language identification.  They converted code-mixed text to native Tamil script by translating English words and transliterating romainzed Tamil words. Similar technique was used by \citet{upadhyay-etal-2021-hopeful} and \citet{srinivasan-2020-msr}.

\section{Methodology}

This shared task was formulated as a multiclass classification problem where the model should be able to predict the existence of any form of homophobia or transphobia in a YouTube comment. The entire pipeline consists of two main components - a classification head on top of different popular models based on the transformer architecture, and a data augmentation technique for oversampling the English dataset. These components have been explained in further detail ahead.

\subsection{Transformer-based Models}

Since its introduction in 2017, the Transformer architecture and its variants have set a new state of the art across several NLP tasks. Various pre-trained language models (PLMs) based on the Transformer architecture were experimented with in this task as mentioned below.

\textbf{BERT} (\verb|bert-base-uncased|) uses the encoder part of the Transformer architecture and has been pretrained on the Book Corpus and English Wikipedia using a masked language modeling (MLM) and next sentence prediction (NSP) objective \citep{devlin2018bert}.

\textbf{mBERT} or multilingual BERT (\verb|bert-base-multilingual-cased|) is a BERT model that has been pretrained on 104 languages across Wikipedia and has shown surprisingly good cross-lingual performance on several NLP tasks.

\textbf{XLM-RoBERTa} (\verb|xlm-roberta-base|) has been pretrained on 2.5TB of massive multilingual data using the MLM objective. It beat mBERT on various cross-lingual benchmarks \citep{conneau2019unsupervised}.

\textbf{IndicBERT} is pretrained on a large-scale corpora of 12 Indian languages. It outperforms mBERT and XLM-RoBERTa on a number of tasks, while having 10 times fewer parameters to train \citep{kakwani2020indicnlpsuite}.

\textbf{HateBERT} is obtained by re-training BERT on RAL-E, a large-scale dataset of reddit comments from banned communities. It outperforms BERT on three English datasets for offensive, abusive language and hate speech detection tasks. \citep{caselli-etal-2021-hatebert}.

\begin{table*}[t]
\centering
\resizebox{\textwidth}{!}{%
\begin{tabular}{@{\extracolsep{6pt}}clllllllll@{}}
\hline
\multicolumn{1}{c}{\multirow{2}{*}{\textbf{Class}}} & \multicolumn{3}{c}{\textbf{English}} & \multicolumn{3}{c}{\textbf{Tamil}} & \multicolumn{3}{c}{\textbf{Tamil-English}}\\
\cline{2-4} \cline{5-7} \cline{8-10}
\multicolumn{1}{c}{} & \textbf{Train} & \textbf{Dev} & \textbf{Test} & \textbf{Train} & \textbf{Dev} & \textbf{Test} & \textbf{Train} & \textbf{Dev} & \textbf{Test} \\
\hline
Homophobic & 157 & 58 & 61 & 485  & 103 & 135 & 311 & 66 & 88 \\
Transphobic & 6 & 2 & 5 & 155 & 37 & 41 & 112 & 38 & 34 \\
Non-anti-LGBT+ content & 3001 & 732 & 924 & 2022 & 526 & 657 & 3438 & 862 & 1085 \\
\hline
\multicolumn{1}{c}{Total} & \multicolumn{3}{c}{4946} & \multicolumn{3}{c}{4161} & \multicolumn{3}{c}{6034}\\
\hline
\end{tabular}}
\caption{\label{dataset-split}
Detailed split of the multilingual dataset of YouTube comments
}
\end{table*}

\subsection{Data Augmentation}

Data augmentation is an important technique to build robust and more generalizable models. There are a number of techniques in NLP, each suitable to a certain task that can be used to augment the data \citep{feng2021survey}. 

For this task (in English), Surface Form Alteration as exhibited by \textit{Easy Data Augmentation} (EDA) was utilized \citep{wei-zou-2019-eda}. EDA produces new data samples by randomly deleting, inserting or swapping the order of words in a sentence. It can also perform synonym replacement for any word selected at random. These four simple, yet effective, operations make EDA easy to use.

\section{Experimental Setup}

In this section we review the setup needed to reproduce the experiments.

\subsection{Datasets} \label{Datasets}

The dataset for the task was provided by the organizers \citep{Chakravarthi2021DatasetFI}. It is a collection of 15,141  multilingual YouTube comments classified as being one of Homophobic, Transphobic, or Non-anti-LGBT+ content. The split of the dataset is shown in Table~\ref{dataset-split}.

\subsection{Preprocessing}

Two different preprocessing methods were adopted. First, punctuation symbols were removed, since social media comments are highly informal and tend to contain large number of punctuation symbols which may dilute the system performance.

In addition, de-emojification was carried out to replace emojis in the text with corresponding English expressions using the Python \verb|emoji| package. Table~\ref{deemoji} displays a sample de-emojification example.

\begin{table}[hbt!]
\centering
\resizebox{\linewidth}{!}{%
\begin{tabular}{ll}
I love it \raisebox{-0.1\height}{\includegraphics[scale=0.04]{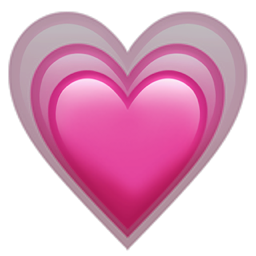}} \raisebox{-0.1\height}{\includegraphics[scale=0.04]{u1F497.png}} \raisebox{-0.1\height}{\includegraphics[scale=0.04]{u1F497.png}}\\
i love it growing heart growing heart growing heart\\
\hline
\end{tabular}}
\caption{Depiction of de-emojification on a sample English YouTube comment}
\label{deemoji}
\end{table}
\subsection{EDA Parameters}

As is visible from Table~\ref{dataset-split}, the dataset is highly imbalanced in its split. The Homophobia class constitutes slightly less than 10\% of the data, while only 2.9\% comments were labeled as being Transphobic. Hence both these classes were subject to oversampling by means of EDA. The class Non-anti-LGBT+ content was downsampled to mitigate the imbalance.

Augmentation was only applied to English comments.

The parameter $\alpha$ (indicating the percent of words in a sentence that are changed) was kept as default ($=0.1$). However the argument $n_{aug}$ (specifying the number of augmentations to be produced for each sample) was chosen to be $16$ and $32$ for Homophobia and Transphobia classes respectively.

\begin{table}[hbt!]
\centering
\resizebox{\linewidth}{!}{%
\begin{tabular}{ll}
\hline
GT & I have to experience like that. So sad\\\hline
RD & i to experience like so sad\\
SR & i have to experience like that so pitiful\\
RI & i have to experience like that distressing so sad\\
RS & experience have to i like that so sad\\\hline
\end{tabular}}
\caption{Depiction of data augmentation on a sample English YouTube comment. GT: ground truth, RD: random deletion, SR: synonym replacement, RI: random insertion, RS: random swapping}
\label{aug}
\end{table}

The final classwise split of the training data is shown in Table~\ref{final-split}.

\begin{table}[hbt!]
\centering
\begin{tabular}{lc}
\hline
\textbf{Class} & \textbf{Final size}\\
\hline
Homophobic & 2826 \\
Transphobic & 204 \\
Non-anti-LGBT+ content & 1500 \\\hline
\end{tabular}
\caption{Classwise split of the training data after EDA augmentation}
\label{final-split}
\end{table}

\subsection{Baseline Methods} \label{Baselines}

We provide baselines for all three tracks based on a simple feature extraction approach.

We use the \verb|[CLS]| token associated with the final hidden state of the transformer model as feature vector for a linear regression classifier.

To extract the hidden state from the checkpoint, we use BERT base model for the English track and mBERT for the other tracks.

\subsection{Setup}

The experiments were run on a Google Colab Pro notebook with Tesla P100 GPU.

For the all tasks, the maximum sequence length was set to $128$ and batch size to $32$. The learning rate and the number of epochs were set to $2e-5$ and $3$ respectively for the English and Tamil track and $3e-5$ and $5$ respectively for the code-mixed track. The choice of EDA parameters was based on suggestions given in the original paper whereas the model hyperparameters were selected based on popular successful configurations.

\section{Results}

The metric used to rank system performances is macro-averaged F1-score. It is calculated as the (unweighted) arithmetic mean of all the per-class F1-scores.

\[
\textrm{Macro-averaged F1-score} =  \frac{1}{N} \sum_{i=1}^{N} F1_i
\]

where $i$ is the class index and $N$ is the number of classes

Tables~\ref{eng-results}, ~\ref{tam-results} and ~\ref{tam-eng-results} list the macro-averaged Precision, macro-averaged Recall and macro-averaged F1-score for various PLMs tested on English, Tamil and code-mixed Tamil-English development dataset respectively.

Similarly Tables~\ref{eng-results-test}, ~\ref{tam-results-test} and ~\ref{tam-eng-results-test} list the corresponding metrics achieved by the final submissions on English, Tamil and Tamil-English test dataset as released by the organizers.

The tables also provide baseline metrics for each track based on the method explained in Section \ref{Baselines}.

\subsection{English}

\begin{table}[hbt!]
\centering
\begin{tabular}{llll}
\hline
\textbf{Model} & \textbf{P} & \textbf{R} & \textbf{F1}\\
\hline
BERT embeddings + LR & 0.40 & 0.47 & 0.42 \\
\hline
BERT base cased & 0.46 & 0.46 & 0.461 \\
XLM-RoBERTa & 0.49 & 0.40 & 0.42 \\
hateBERT & 0.50 & 0.44 & 0.461 \\
mBERT & 0.48 & 0.45 & \textbf{0.462} \\\hline
\end{tabular}
\caption{Performance of various PLMs on augmented, preprocessed English development dataset}
\label{eng-results}
\end{table}

\begin{table}[hbt!]
\centering
\begin{tabular}{llll}
\hline
\textbf{Model} & \textbf{P} & \textbf{R} & \textbf{F1}\\
\hline
mBERT & 0.43 & 0.42 & 0.42 \\\hline
\end{tabular}
\caption{Performance of best peforming system (\textit{mBERT}) on preprocessed English test dataset}
\label{eng-results-test}
\end{table}

\subsection{Tamil}

Here we investigate the performance of some popular multilingual models that were trained on Tamil language.

\begin{table}[hbt!]
\centering
\begin{tabular}{llll}
\hline
\textbf{Model} & \textbf{P} & \textbf{R} & \textbf{F1}\\
\hline
mBERT embeddings + LR & 0.71 & 0.59 & 0.63\\
\hline
IndicBERT & 0.48 & 0.47 & 0.47 \\
XLM-RoBERTa & 0.47 & 0.55 & 0.50 \\
mBERT & 0.77 & 0.71 & \textbf{0.72} \\\hline
\end{tabular}
\caption{Performance of various PLMs on preprocessed Tamil development dataset}
\label{tam-results}
\end{table}

\begin{table}[hbt!]
\centering
\begin{tabular}{llll}
\hline
\textbf{Model} & \textbf{P} & \textbf{R} & \textbf{F1}\\
\hline
mBERT & 0.69 & 0.61 & 0.64 \\\hline
\end{tabular}
\caption{Performance of best peforming system (\textit{mBERT}) on preprocessed Tamil test dataset}
\label{tam-results-test}
\end{table}

\subsection{Tamil-English}

For the code-mixed task, we analyze the performance of the same set of multilingual models that were experimented with on the Tamil task.

\begin{table}[hbt!]
\centering
\begin{tabular}{llll}
\hline
\textbf{Model} & \textbf{P} & \textbf{R} & \textbf{F1}\\
\hline
mBERT embeddings + LR & 0.61 & 0.47 & 0.51\\
\hline
IndicBERT & 0.39 & 0.41 & 0.40 \\
XLM-RoBERTa & 0.40 & 0.43 & 0.41 \\
mBERT & 0.67 & 0.52 & \textbf{0.54} \\\hline
\end{tabular}
\caption{Performance of various PLMs on preprocessed Tamil-English development dataset}
\label{tam-eng-results}
\end{table}

\begin{table}[hbt!]
\centering
\begin{tabular}{llll}
\hline
\textbf{Model} & \textbf{P} & \textbf{R} & \textbf{F1}\\
\hline
mBERT & 0.61 & 0.56 & 0.58 \\\hline
\end{tabular}
\caption{Performance of best peforming system (\textit{mBERT}) on preprocessed Tamil-English test dataset}
\label{tam-eng-results-test}
\end{table}

\section{Ablation Study}

In this section we discuss the effect of preprocessing and data augmentation (DA) on the model performance.

The dataset as described in Section \ref{Baselines} is highly skewed towards the \textit{Non-anti-LGBT+ content} class. Hence it makes sense to compare the performance of a majority classifier with that of the models submitted for evaluation.

We train a dummy classifier based on most-frequent strategy and tabulate the results (macro-averaged Precision, Recall and F1-score) in Table~\ref{dummy}. We deliberately use the un-augmented version of preprocessed English dataset to show the show the performance of the majority classifier without handling class imbalance.

\begin{table}[hbt!]
\centering
\begin{tabular}{llll}
\hline
\textbf{} & \textbf{P} & \textbf{R} & \textbf{F1}\\
\hline
English & 0.31 & 0.33 & 0.32 \\
Tamil & 0.26 & 0.33 & 0.29 \\
Code-mixed & 0.30 & 0.33 & 0.31 \\\hline
\end{tabular}
\caption{Performance of dummy majority classifier on the dataset}
\label{dummy}
\end{table}

The poor performance is a consequence of the extreme class imbalance which we aim to solve by data augmentation. However, not all DA techniques prove to be effective for all NLP tasks. Thus we also analyze the effect of preprocessing and DA on the performance of transformer models.

Table~\ref{ablation} analyzes the efficacy of EDA as a DA technique for handling class imbalance in our English dataset. It also divides a line between the performance of the model on the stock dataset v/s one that has been preprocessed.

\begin{table}[hbt!]
\centering
\resizebox{\linewidth}{!}{%
\begin{tabular}{lllllc}
\hline
 & \textbf{Setting} & \textbf{P} & \textbf{R} & \textbf{F1} & \textbf{Rel. }\\
\hline
\multirow{3}{*}{English} & base & 0.52 & 0.40 & 0.43 &  \\
 & +PRE & 0.40 & 0.43 & 0.41 & $\downarrow$ \\
 & +DA & 0.52 & 0.37 & 0.39 & $\downarrow$ \\\hline
 \multirow{3}{*}{Tamil} & base & 0.73 & 0.75 & 0.74 \\
 & +PRE & 0.70 & 0.73 & 0.72 & $\downarrow$ \\\hline
 \multirow{3}{*}{Code-mixed} & base & 0.43 & 0.42 & 0.43 \\
 & +PRE & 0.71 & 0.56 & 0.60 & $\uparrow$ \\\hline
\end{tabular}}
\caption{Performance of mBERT on the stock version of the dataset as it is (base), preprocessed dataset (+PRE) and augmented but non-preprocessed English dataset (+DA)}
\label{ablation}
\end{table}

We observe that preprocessing (de-emojification in all three tracks and de-punctuation in the case of only English) does not increase the macro-averaged F1 score for English and Tamil. Infact it reduces the score by a small margin. However, we notice a significant improvement in the case of code-mixing.

We also observe that EDA is not an efficient DA technique as it fails to handle the class imbalance. Transformer models were able to successfully predict with higher precision  and recall in the absence of any augmentation and with limited samples.

\section{Conclusion and Future Work}

Homophobia and transphobia have not been the focus of many umbrella hate speech detection tasks. We examined the ability of pretrained large transformer-based models to detect homophobia and transphobia in a corpus of YouTube comments written in English and Tamil. Experimental results demonstrated that multilingual BERT performed the best on both language tasks, and the code-mixed task as well, without being exposed to any code-mixing beforehand. This can be attributed to its capability for zero-shot cross-lingual transfer when fine-tuned on downstream tasks.

From Section \hyperlink{section.6}{6} we also observed that the effect of preprocessing was largely dependent on the choice of language setting. This makes sense considering the difference in underlying language constructs. Tamil, for instance, does not make use of standard English-based punctuation marks. On the other hand, we conclude that the choice of an effective DA techniqe depends on the underlying task and the data source. Social media data often lacks linguistic purism and hence, token perturbations such as those introduced by EDA did not help.

In the future, we would like to adopt a more aggressive DA technique such as that involving text generation (text In-filling, generating typos) or an auxilliary dataset (kNN, LM decoding). We would also like to evaluate the effect of translation and transliteration on code-mixed text classification.

\section*{Acknowledgments}

We would like to acknowledge the efforts of the workshop organizers in effecting positive social change through AI by conducting such shared tasks. We also thank the reviewers for their time and insightful comments.

\bibliography{anthology,custom}
\bibliographystyle{acl_natbib}

\appendix

\end{document}